\pdfoutput=1

\documentclass[11pt]{article}

\usepackage{emnlp2021}

\usepackage{times}
\usepackage{latexsym}
\usepackage{amsmath}
\usepackage{multirow}
\usepackage{graphicx}

\usepackage[T1]{fontenc}

\usepackage[utf8]{inputenc}
\usepackage{color}

\usepackage{microtype}

\title{Unsupervised Keyphrase Extraction by Jointly Modeling Local and Global Context}
\author{Xinnian Liang\textsuperscript{1}\footnotemark[1], Shuangzhi Wu\textsuperscript{2}, Mu Li\textsuperscript{2} \and Zhoujun Li\textsuperscript{1}\footnotemark[2]\\ 
  \textsuperscript{1}State Key Lab of Software Development Environment, Beihang University, Beijing, China \\
  \textsuperscript{2}Tencent Cloud Xiaowei, Beijing, China\\
  \texttt{\{xnliang,lizj\}@buaa.edu.cn};
  \texttt{\{frostwu,ethanlli\}@tencent.com};\\
}

\begin{document}
\maketitle

\renewcommand{\thefootnote}{\fnsymbol{footnote}}
\footnotetext[1]{Contribution during internship at Tencent Inc.} 
\footnotetext[2]{Corresponding Author}

\begin{abstract}
Embedding based methods are widely used for unsupervised keyphrase extraction (UKE) tasks. Generally, these methods simply calculate similarities between phrase embeddings and document embedding, which is insufficient to capture different context for a more effective UKE model. In this paper, we propose a novel method for UKE, where local and global contexts are jointly modeled. From a global view, we calculate the similarity between a certain phrase and the whole document in the vector space as transitional embedding based models do. In terms of the local view, we first build a graph structure based on the document where phrases are regarded as vertices and the edges are similarities between vertices. Then, we proposed a new centrality computation method to capture local salient information based on the graph structure. Finally, we further combine the modeling of global and local context for ranking. We evaluate our models on three public benchmarks (Inspec, DUC 2001, SemEval 2010) and compare with existing state-of-the-art models. The results show that our model outperforms most models while generalizing better on input documents with different domains and length. Additional ablation study shows that both the local and global information is crucial for unsupervised keyphrase extraction tasks.
\end{abstract}

\section{Introduction}
Keyphrase extraction (KE) task aims to extract a set of words or phrases from a document that can represent the salient information of the document \cite{hasan-ng-2014-automatic}. KE models can be divided into supervised and unsupervised.
Supervised methods need large-scale annotated training data and always perform poorly when transferred to different domain or type datasets.
Compared with the supervised method, the unsupervised method is more universal and adaptive via extracting phrases based on information from input document itself.
In this paper, we focus on the unsupervised keyphrase extraction (UKE) model.

UKE has been widely studied \cite{mihalcea-2004-graph,wan-2008-single,bougouin-etal-2013-topicrank,boudin-2018-unsupervised,bennani-smires-etal-2018-simple,ieee-sun-etal-2020-sifrank} in the keyphrase extraction field. Recently, with the development of text representation, embedding-based models \cite{bennani-smires-etal-2018-simple,ieee-sun-etal-2020-sifrank} have achieved promising results and become the new state-of-the-art models. Usually, these methods compute phrase embeddings and document embedding with static word2vec models (e.g. GloVe \cite{pennington-etal-2014-glove,10.5555/3044805.3045025,pgj2017unsup}) or dynamic pre-trained language models (e.g. BERT \cite{devlin2019bert}). Then, they rank candidate phrases by computing the similarity between phrases and the whole document in the vector space. Though, these methods performed better than traditional methods \cite{mihalcea-2004-graph,wan-2008-single,bougouin-etal-2013-topicrank}, the simple similarity between phrase and document is insufficient to capture different kinds of context and limits in performance. 

\begin{figure}
    \centering
    \includegraphics[scale=0.7]{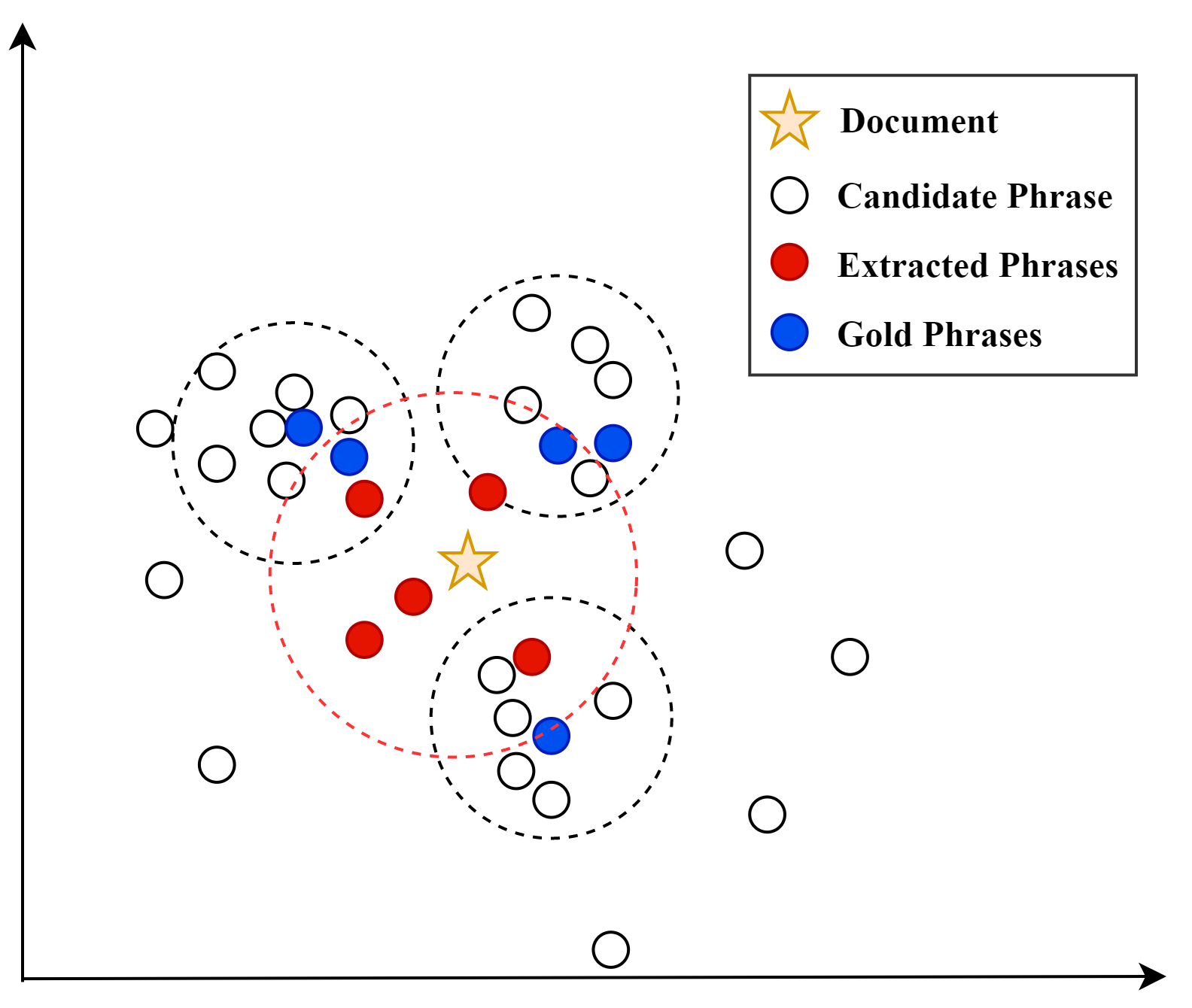}
    \caption{Visualization of embedding space. Nodes refer to candidate phrase representation and star is document representation. Black circles mean clusters which contain local salient information. Red circle means global similarity phrases.}
    \label{fig:example}
\end{figure}
Figure \ref{fig:example} shows an intuitive explanation for the importance of context modeling.
The nodes are candidate phrase embeddings, the star is the document embedding. 
Each black circle represents one local context.
Nodes in the same black circle mean that these candidate phrases are all related to one vital local information (e.g. one topic/aspect of the document).
Nodes in the red circle mean that these candidate phrases are similar with the document semantics. 
If only model the global context via computing similarity between candidate phrases and the document, the model will tend to select red nodes, which will ignore local salient information in three clusters.
In order to get the keywords accurately, we should take the local context (black circles) and global context (red circle) into consideration.

To obtain information from context adequately, in this paper, we proposed a novel method which jointly models the local and global context of the input document. 
Specifically, we calculate the similarity between candidate phrases and the whole document for modeling global context.
For local context modeling, we first build a graph structure, which represents each phrase as nodes and the edges are similarity between nodes.
Then, we proposed a new centrality computation method, which is based on the insight that the most important information typically occurs at the start or end of documents (document boundary) \cite{lin-hovy-1997-identifying-bd1,teufel-1997-sentence-bd2,dong-2021-discourseaware-bd3}, to measure salience of local context based on the graph structure. 
Finally, we further combine the measure of global similarity and local salience for ranking. 
To evaluate the effectiveness of our method, we compare our method with recent state-of-the-art models on three public benchmarks (Inspec, DUC 2001, SemEval 2010). 
The results show that our model can outperform most models while generalizing better on input documents with different domains and length. It is deservedly mentioned that our models have a huge improvement on long scientific documents. 

\begin{figure*}[]
    \centering
    \includegraphics[scale=0.65]{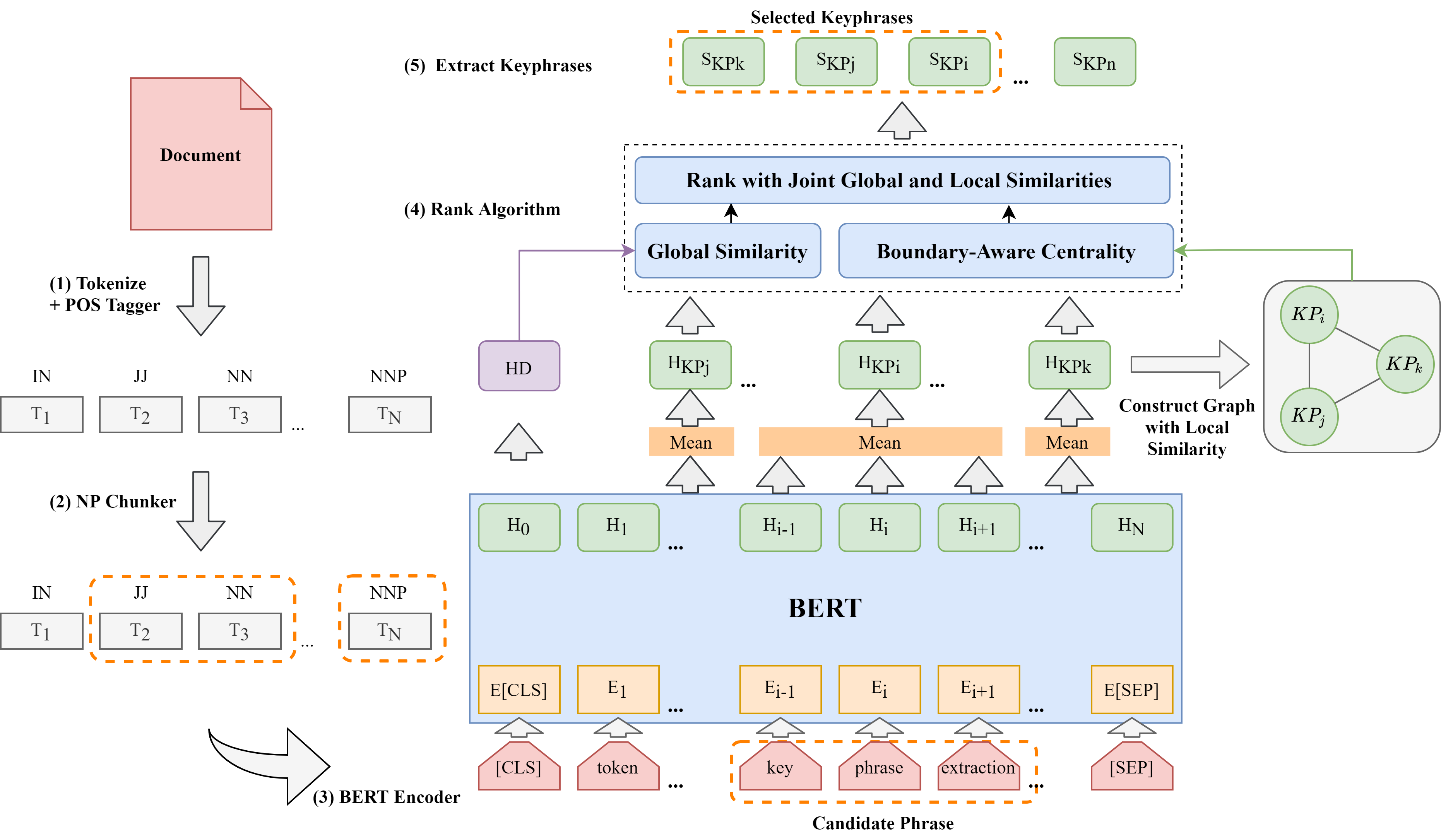}
    \caption{The framework of our unsupervised keyphrase extraction ranking model. (1) Get tokenized document and POS tags. (2) Extract noun phrases that consist of zero or more adjectives followed by one or multiple nouns. (3) Obtain embeddings of tokens in document with BERT. (4) Compute boundary-aware centrality and global relevance of each candidate phrases with global and local similarities. (5) Rank and extract keyphrases from candidate phrases with scores from the previous step.}
    \label{fig:framework}
\end{figure*}

\section{Methodology}
The overall framework of our model is shown in Fig. \ref{fig:framework}. We follow the general process of unsupervised keyphrase extraction. The main steps are as follows: (1) We tokenize the document and tag the document with part-of-speech (POS) tags. (2) We extract candidate phrases based on part-of-speech tags. We only keep noun phrases (NP) that consist of zero or more adjectives followed by one or multiple nouns \cite{wan-and-xiao-2008-npcandidate}. (3) We use a pre-trained language model to map the document text to low-dimension vector space and extract vector representation of candidate phrases and the whole document. (4) We score each candidate phrase with a rank algorithm which jointly models the global and local context. (5) We extract phrases with scores from the rank algorithm.

The main contribution of the whole process is the rank algorithm we proposed in step (4), which can be divided into three components: 1) phrase-document similarity for modeling global context; 2) boundary-aware centrality for modeling local context; 3) the combination of global and local information. We will introduce the details of these components in this section.

\subsection{Document and Phrases Representations}
Before introducing the rank algorithm, we first make clear step (1) - (3).
We follow the common practice and use \textit{StanfordCoreNLP Tools}\footnote{https://stanfordnlp.github.io/CoreNLP/} to accomplish step (1) and (2).
After previous universal steps, the document $D$ was tokenized into tokens $\{t_1, t_2, ..., t_N\}$ and candidate phrases $\{KP_0, KP_1, ..., KP_n\}$ were extracted from document $D$.
Different from previous works \cite{bennani-smires-etal-2018-simple} which use static vector to represent tokens in document, we employ BERT, which is a strong pre-trained language model, to obtain contextualized dynamic vector representations by Equ. (1).
\begin{equation}
    \{H_1, H_2, ..., H_N\} = \mathtt{BERT}(\{t_1, t_2, ..., t_N\})
\end{equation}
Where $H_i$ is the vector representation of token $t_i$. Then, we obtain the vector representation $H_{\mathtt{KP}_i}$ of candidate phrases by computing the average of the phrase's token vectors. 
The document vector representation is computed with max-pooling operation by Equ. (2).
\begin{equation}
    H_D=\mathtt{Maxpooling}(\{H_1, H_2, ..., H_N\})
\end{equation}
Then, we can obtain document vector representation $H_D$ which contains the global semantic information of the full document and a set of candidate phrase representations $V=\{H_{\mathtt{KP}_i}\}_{i=1,\dots,n}$. Based on these representations, we will introduce the core rank algorithm of our model in the next section.

\subsection{Proposed Rank Algorithm}
\subsubsection{Phrase-Document Similarity}
We first introduce the computation of the phrase-document similarity for modeling global context.
Specifically, we empirically employ Manhattan Distant (i.e. L1-distance) to compute similarity by Equ. (3).
\begin{equation}
    \mathbf R(H_{\mathtt{KP}_i}) = \frac{1}{\parallel H_D - H_{\mathtt{KP}_i} \parallel_1}
\end{equation}
Where $\parallel \cdot \parallel_1$ means Manhattan Distant and $R(H_{\mathtt{KP}_i})$ represent the relevance between candidate phrase $i$ and the whole document.

\subsubsection{Traditional Degree Centrality}
Graph-based ranking algorithms for keyphrase extraction represent a document as a graph $\mathcal G=(V, E)$, where $V=\{H_{\mathtt{KP}_i}\}_{i=1,\dots,n}$ is the set of vector that represent nodes in graph (i.e. candidate phrases in document), and $E=\{e_{ij}\}$ is the set of edges that represent interactions between candidate phrases. 
In this paper, we simply employ the degree of nodes as centrality to measure the importance of nodes. The degree centrality for candidate phrase $i$ can be computed with Equ. (4).
\begin{equation}
    \mathbf C(H_{\mathtt{KP}_i}) = \sum_{j=1}^n e_{ij}
\end{equation}
Where $e_{ij} = H_{\mathtt{KP}_i}^T\cdot H_{\mathtt{KP}_j}$ is the dot-product similarity score for each pair $(H_{\mathtt{KP}_i}, H_{\mathtt{KP}_j})$. We could also use other similarity measure methods (e.g. cosine similarity), but we empirically find that the simple dot-product performs better.

\subsubsection{Boundary-Aware Centrality}
Traditional centrality computation is based on the assumption that the contribution of the candidate phrase's importance in the document is not affected by the relative position of them, and the similarities of two graph nodes are symmetric. From human intuition, phrases that exist at the start or the end of a document should be more important than others. 
To implement this insight, we propose a new centrality computation method called boundary-aware centrality based on the assumption that important information typically occurs near boundaries (the start and end of documents) \cite{lin-hovy-1997-identifying-bd1,teufel-1997-sentence-bd2}. 

We reflect this assumption by employing a boundary function $d_b(i)$ over position of candidate phrases. This function $d_b$ is formulated as Equ (5).
\begin{equation}
    d_b(i) = \min(i, \alpha (n-i))
\end{equation}
Where $n$ is the number of candidate phrases, and $\alpha$ is a hyper-parameter that controls relative importance of the start and end of a document. 
For node $i$ and $j$, if $d_b(i) < d_b(j)$, then node $i$ is closer to the boundary than node $j$. When calculating the centrality of node $i$, we need to reduce the contribution of node $j$ to the centrality of node $i$.
Based on this assumption and the boundary function $d_b(i)$, we can reconstruct the centrality computation of node $i$ in the graph as Equ. (6).
\begin{equation}
    \mathbf C(H_{\mathtt{KP}_i}) = \sum_{d_b(i) < d_b(j)}e_{ij} + \lambda\sum_{d_b(i) \geq d_b(j)}e_{ij}
\end{equation}
Where $\lambda$ is used to reduce the influence of phrases which do not appear near the boundary to the centrality of node $i$.

Besides, we employ a threshold $\theta = \beta (\max(e_{ij} - \min(e_{ij}))$ to filter the noise from nodes, which is far different from node $i$. We remove the influence of them to centrality by setting all $e_{ij} < \theta$ to zero. $\beta$ is a hyper-parameter that controls the filter boundary. With the introduction of the noise filter strategy, we rewrite the Equ. (6) as Equ. (7).
\begin{equation}
    \begin{aligned}
        \mathbf C(H_{\mathtt{KP}_i}) =& \sum_{d_b(i) < d_b(j)}\max(e_{ij} - \theta, 0) \\
        +& \lambda\sum_{d_b(i) \geq d_b(j)}\max(e_{ij} - \theta, 0)
    \end{aligned}
\end{equation}
Where $\mathbf C(H_{\mathtt{KP}_i})$ represents the local salience of candidate phrase $i$.

For most long documents or news articles, the author tends to write the key information at the beginning of the document.
\citet{AAAI1714377} point out that the position-biased weight can greatly improve the performance for keyphrase extraction and they employ the sum of the position's inverse of words in the document as the weight. For example, the word appearing at 2th, 5th and 10th, has a weight $p(w_i) = 1/2 + 1/5 + 1/10 = 0.8$.
Our boundary-aware centrality has considered relative position information with boundary function. To prevent double counting, we follow a simpler position bias weight from \cite{ieee-sun-etal-2020-sifrank}, which only considers where the candidate phrase first appears. The position bias weight is computed by $p(\mathtt{KP}_i) = \frac{1}{p_1}$, where $p_1$ is the position of the candidate keyphrase's first appearance. After that the softmax function is used to normalize the position bias weight as follow:
\begin{equation}
    \hat{p}(\mathtt{KP}_i) =  \frac{\exp(p(\mathtt{KP}_i))}{\sum_{k=1}^n\exp(p(\mathtt{KP}_k))}
\end{equation}
Then, boundary-aware centrality can be rewritten as Equ. (9).
\begin{equation}
    \mathbf{\hat{C}}({H_{\mathtt{KP}_i}}) = \hat{p}(\mathtt{KP}_i) \cdot \mathbf C(H_{\mathtt{KP}_i})
\end{equation}
We finally employ $\mathbf{\hat{C}}({H_{\mathtt{KP}_i}})$ to measure the local salience of candidate phrase $i$.

\subsubsection{Rank with Global and Local Information}
To consider global and local level information at the same time, we simply combine the measure of global relevance $\mathbf R(H_{\mathtt{KP}_i})$ and local salience $\mathbf{\hat{C}}({H_{\mathtt{KP}_i}})$ of candidate phrase together with multiplication to obtain the final score by Equ. (10).
\begin{equation}
    \mathbf S({H_{\mathtt{KP}_i}}) = \mathbf R(H_{\mathtt{KP}_i}) \cdot \mathbf{\hat{C}}({H_{\mathtt{KP}_i}})
\end{equation}
Finally, we rank candidate phrases with their final score $S({H_{\mathtt{KP}_i}})$ and extract top-ranked $k$ phrases as keyphrases of the document.

\section{Experiments}
\begin{table*}[ht]
\centering
\setlength\tabcolsep{3pt}
\begin{tabular}{|l|ccc|ccc|ccc|}
\hline
\multirow{2}{*}{\textbf{Models}} & \multicolumn{3}{c|}{\textbf{DUC2001}} & \multicolumn{3}{c|}{\textbf{Inspec}} & \multicolumn{3}{c|}{\textbf{SemEval2010}} \\ 
 & F1@5 & F1@10 & F1@15 & F1@5 & F1@10 & F1@15 & F1@5 & F1@10 & F1@15 \\ \hline
\multicolumn{10}{|c|}{Statistical Models} \\ \hline
TF-IDF & 9.21 & 10.63 & 11.06 & 11.28 & 13.88 & 13.83 & 2.81 & 3.48 & 3.91 \\
YAKE & 12.27 & 14.37 & 14.76 & 18.08 & 19.62 & 20.11 & 11.76 & 14.4 & 15.19 \\ \hline
\multicolumn{10}{|c|}{Graph-based Models} \\ \hline
TextRank & 11.80 & 18.28 & 20.22 & 27.04 & 25.08 & 36.65 & 3.80 & 5.38 & 7.65 \\
SingleRank & 20.43 & 25.59 & 25.70 & 27.79 & 34.46 & 36.05 & 5.90 & 9.02 & 10.58 \\
TopicRank & 21.56 & 23.12 & 20.87 & 25.38 & 28.46 & 29.49 & 12.12 & 12.90 & 13.54 \\
PositionRank & 23.35 & 28.57 & 28.60 & 28.12 & 32.87 & 33.32 & 9.84 & 13.34 & 14.33 \\
MultipartiteRank & 23.20 & 25.00 & 25.24 & 25.96 & 29.57 & 30.85 & 12.13 & 13.79 & 14.92 \\ \hline
\multicolumn{10}{|c|}{Embedding-based Models} \\ \hline
EmbedRank d2v & 24.02 & 28.12 & 28.82 & 31.51 & 37.94 & 37.96 & 3.02 & 5.08 & 7.23 \\
EmbedRank s2v & 27.16 & 31.85 & 31.52 & 29.88 & 37.09 & 38.40 & 5.40 & 8.91 & 10.06 \\
SIFRank & 24.27 & 27.43 & 27.86 & 29.11 & 38.80 & 39.59 & - & - & - \\
SIFRank+ & \textbf{30.88} & 33.37 & 32.24 & 28.49 & 36.77 & 38.82 & - & - & - \\
KeyGames & 24.42 & 28.28 & 29.77 & 32.12 & \textbf{40.48} & 40.94 & 11.93 & 14.35 & 14.62 \\ \hline
\multicolumn{10}{|c|}{Proposed Model} \\ \hline
Our Model & 28.62 & \textbf{35.52} & \textbf{36.29} & \textbf{32.61} & 40.17 & \textbf{41.09} & \textbf{13.02} & \textbf{19.35} & \textbf{21.72} \\ \hline
\end{tabular}
\caption{Comparison of our models with other baselines.}
\label{tab:res}
\end{table*}

\begin{table*}[]
\centering
\setlength\tabcolsep{3pt}
\begin{tabular}{|l|ccc|ccc|ccc|}
\hline
\multirow{2}{*}{} & \multicolumn{3}{c|}{\textbf{DUC2001}} & \multicolumn{3}{c|}{\textbf{Inspec}} & \multicolumn{3}{c|}{\textbf{SemEval2010}} \\
 & F1@5 & F1@10 & F1@15 & F1@5 & F1@10 & F1@15 & F1@5 & F1@10 & F1@15 \\ \hline
Our Model & 28.62 & 35.52 & 36.29 & 32.49 & 40.04 & 41.05 & 12.26 & 19.22 & 21.42 \\ \hline
- Global Similarity & 26.22 & 33.85 & 34.61 & 30.75 & 38.49 & 40.52 & 12.00 & 18.93 & 21.29 \\
- Local Similarity & 17.45 & 18.64 & 19.03 & 22.51 & 27.58 & 30.36 & 8.81 & 10.7 & 11.45 \\ \hline
\end{tabular}
\caption{The results of ablation experiments on three datasets.}
\label{tab:abl}
\end{table*}

\subsection{Datasets and Evaluation Metrics}
We evaluate our model on three public datasets: \textbf{Inspec}, \textbf{DUC2001} and \textbf{SemEval2010}. The \textbf{Inspec} dataset \cite{hulth-2003-improved} consists of 2,000 short documents from scientific journal abstracts. We follow previous works \cite{bennani-smires-etal-2018-simple,ieee-sun-etal-2020-sifrank} to use 500 test documents and the version of uncontrolled annotated keyphrases as ground truth. 
The \textbf{DUC2001} dataset \cite{wan-2008-single} is a collection of 308 long length news articles with average 828.4 tokens.
The \textbf{SemEval2010} dataset \cite{kim-etal-2010-semeval} contains ACM full length papers. In our experiments, we use the 100 test documents and the combined set of author- and reader- annotated keyphrases.

We follow the common practice and evaluate the performance of our models in terms of f-measure at the top N keyphrases (F1@N), and apply stemming to both extracted keyphrases and gold truth. Specifically, we report F1@5, F1@10 and F1@15 of each model on three datasets.

\subsection{Comparison Models and Implementation Details}
We compare our methods with three types of models to comprehensively prove the effectiveness of our models.
Firstly, we compare with traditional statistical methods TF-IDF and YAKE \cite{yake-2018}.
Secondly, We compare five strong graph-based ranking methods. TextRank \cite{mihalcea-tarau-2004-textrank} is the first attempt to convert text to graph with the co-occurrence of words and employ PageRank to rank phrases. SingleRank \cite{wan-2008-single} improves the graph construction with a slide window. TopicRank \cite{bougouin-etal-2013-topicrank} considers keyphrase extraction with topic distribution. PositionRank \cite{florescu-caragea-2017-positionrank} employs position information to weight the importance of phrases. MultipartiteRank \cite{boudin-2018-unsupervised} splits the whole graph into sub-graph and ranks them with some graph theory.
Finally, We compare three state-of-the-art embedding-based models. EmbedRank \cite{bennani-smires-etal-2018-simple} first employs embedding of texts with Doc2Vec/Sent2Vec and measures the relevance of phrases and documents to select keyphrases. SIFRank \cite{ieee-sun-etal-2020-sifrank} improves EmbedRank with contextualized embedding from a pre-trained language model. KeyGames \cite{saxena-etal-2020-keygames} creatively introduces game theoretic approach into automatic keyphrase extraction.

All the models use \textit{Stanford CoreNLP Tools}\footnote{https://stanfordnlp.github.io/CoreNLP/} for tokenizing, part-of-speech tagging and noun phrase chunking. And regular expression $\{ \langle NN.\ast | JJ\rangle \ast \langle NN. \ast \rangle\}$ is used to extract noun phrases as the candidate keyphrases.
Our model’s hyperparameters for testing are chosen based on our results with the sampled 200 validation sets.
The test results are chosen from the following hyper-parameter settings: $\alpha \in \{0.5, 0.8, 1, 1.2, 1.5\}$, $\beta \in \{0.0, 0.1, 0.2, 0.3\}$ and $\lambda \in \{0.8, 0.9, 1.0\}$.

\subsection{Results}

We report the results of our model in Tab. \ref{tab:res}. 
We can observe that our model consistently outperforms most of the existing systems across the three datasets, each with different document length, covering two different domains. 
SIFRank and SIFRank+ have a remarkable performance on datasets with short input length due to document embedding of short documents can better represent the semantic information of full document and short document has fewer local information (e.g. aspects), which make embedding-based models perform well. 
We can further see that models with global similarity (i.e. EmbedRank and SIFRank) all outperform graph-based models on short length documents (i.e. DUC2001 and Inspec).

Compared with other works, our model and KeyGames, which is based on game theory, are more generalized and can tackle short and long input documents well.
The advantages of our models are very obvious on the long scientific document dataset SemEval2010. This mainly benefits from the boundary-aware centrality for modeling local context.
\subsection{Discussion}
\subsubsection{Ablation Study}

We evaluate the contribution of the global and local component of our model with ablation study and the results can be seen in Tab. \ref{tab:abl}. 
From the results, we can find that the modeling of local context is more important than the modeling of global context. When we remove local information from our model, our model goes back to an embedding-based model.
The performance on SemEval2010 is not sensitive to the removal of relevance-aware weighting. We guess that embedding of long documents may contain multi-aspects information which influences the measure of similarity between the phrase and the whole document, which leads to the influence of global information being limited.
Overall, we can prove that jointly modeling global and local context is crucial for unsupervised keyphrase extraction and the revisit of degree centrality is effective for modeling local context and meaningful for future work.

\subsubsection{Impact of Hyper-Parameters}
\begin{figure}[ht]
    \centering
    \includegraphics[scale=0.45]{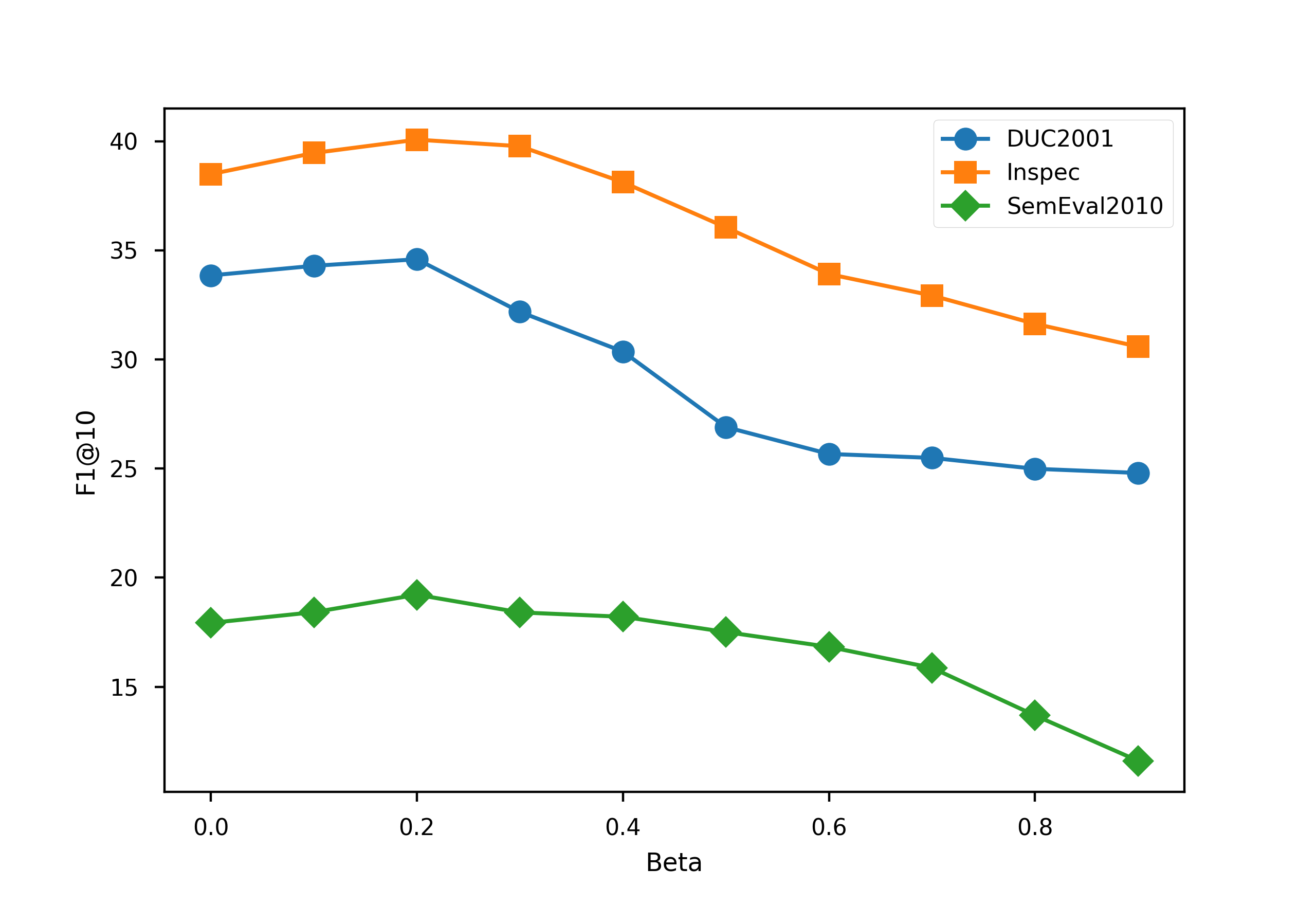}
    \caption{Performance with different $\beta$.}
    \label{fig:beta}
\end{figure}
In this section, we first analyze the impact of hyper-parameter $\beta$ and then report the best setting on each dataset.
We employ three hyper-parameters in our models, $\beta$ is used to filter noise and we can see the impact of $\beta$ from Fig. \ref{fig:beta}. $\beta=0.2$ is a proper configure for three datasets.
$\alpha$ is used to control the importance of the start or end of a document. $\alpha < 1$ means the start of the document is more vital and $\alpha > 1$ means the end of the document is more vital.

The best settings are $\alpha=0.8, \beta=0.2, \lambda=0.9$ on DUC2001, $\alpha=0.5, \beta=0.2, \lambda=0.9$ on Inspec and $\alpha=1.5, \beta=0.2, \lambda=0.8$ on SemEval2010. 
From these settings, we can get the following three conclusions, which is conforming to the characteristics of these datasets.
For DUC2001 and Inspec, most vital information occurs at the start of the document due to the fact that DUC2001 is from news articles, and Inspec is from the abstract.
For SemEval2010, the setting of $\alpha$ is contrary to previous datasets due to SemEval2010 is long scientific documents and much key information occurring at the end of the document (section conclusion). 
The settings of $\lambda$ on three datasets show that long documents need to reduce more influence from contexts not near the boundary, which is intuitive.

\begin{table*}[ht]
\centering
\setlength\tabcolsep{3pt}
\begin{tabular}{|c|ccc|ccc|ccc|}
\hline
\multirow{2}{*}{\textbf{Similarity Measure}} & \multicolumn{3}{c|}{\textbf{DUC2001}} & \multicolumn{3}{c|}{\textbf{Inspec}} & \multicolumn{3}{c|}{\textbf{SemEval2010}} \\
 & F1@5 & F1@10 & F1@15 & F1@5 & F1@10 & F1@15 & F1@5 & F1@10 & F1@15 \\ \hline
Euclidean Distance & 23.31 & 28.04 & 30.39 & 28.5 & 37.01 & 29.25 & 10.99 & 16.37 & 18.41 \\
Cosine similarity & 15.01 & 17.96 & 19.44 & 23.67 & 30.26 & 33.35 & 9.70 & 12.22 & 13.29 \\ 
Manhattan Distance & \textbf{28.62} & \textbf{35.52} & \textbf{36.29} & \textbf{32.49} & \textbf{40.04} & \textbf{41.05} & \textbf{12.26} & \textbf{19.22} & \textbf{21.42} \\\hline
\end{tabular}
\caption{The results of different measure methods for similarity between candidate phrase and the whole document.}
\label{tab:rel}
\end{table*}

\subsubsection{Impact of Different Similarity Measure Methods}

Our model employs Manhattan Distance to measure the similarity between phrases and the whole document. We also attempt to employ different measure methods. The results of different similarity measure methods are shown in Tab. \ref{tab:rel}, and we can see that the advantage of Manhattan Distance is obvious. We also can see that cosine similarity performs badly and is not suitable for our models.

\subsubsection{Case Study}
\begin{figure*}[ht]
    \centering
    \includegraphics[scale=0.36]{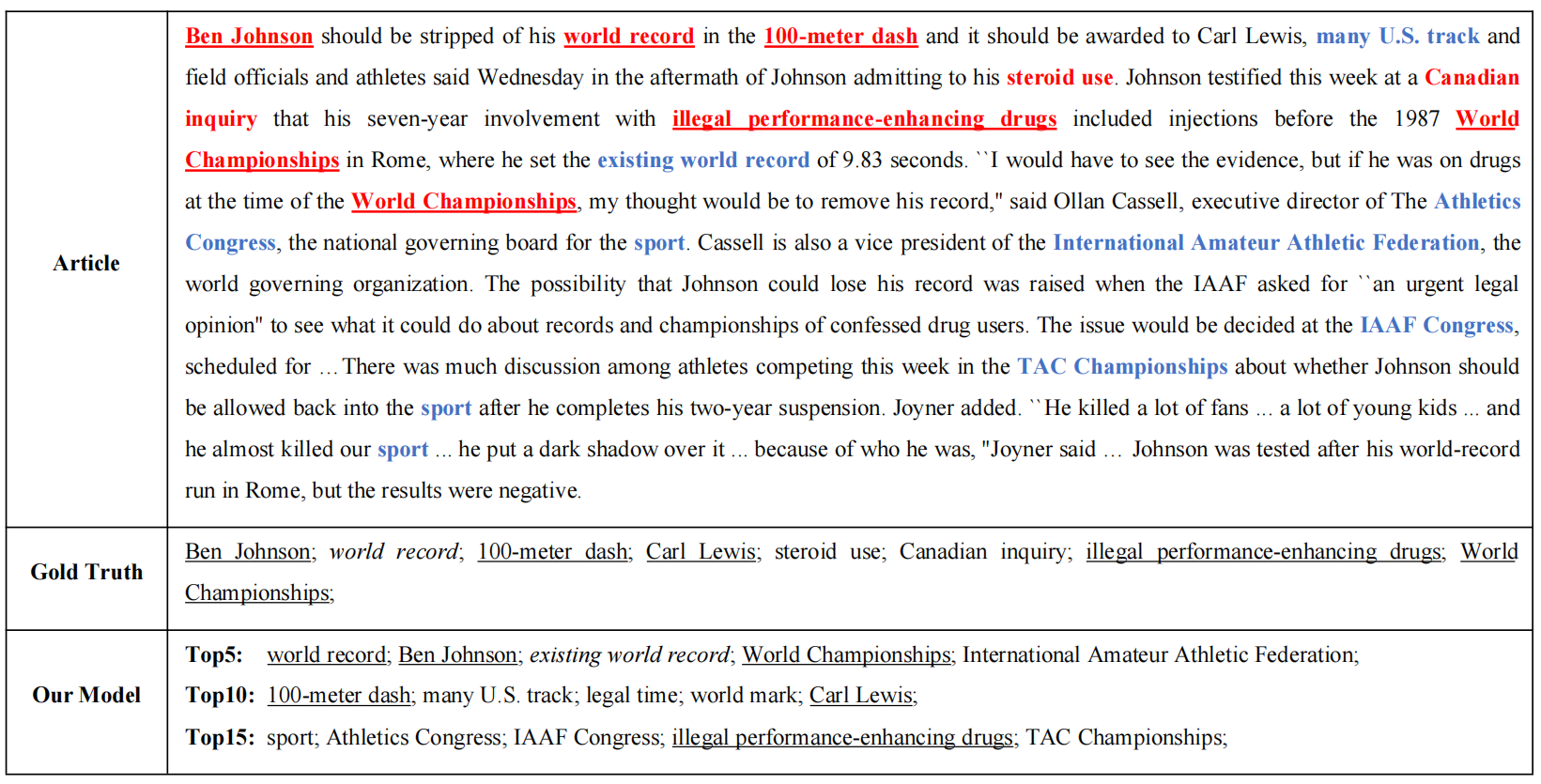}
    \caption{An example from DUC2001. The correct keyphrases are underlined. Red text means the extracted gold truth and blue text means extracted phrases by our model.}
    \label{fig:case}
\end{figure*}
In this section, we show an example from DUC2001 in Fig. \ref{fig:case}. DUC2001 is a dataset from news articles. 
The correct keyphrases are underlined. Red text means the extracted gold truth and blue text means extracted phrases by our model. 

We can see that all keyphrases occur at the start of the document. Our model extracted many correct phrases which are the same as gold truth and extracted the phrase ``existing word record" which is semantically same with ``word record" in gold truth. It is worth mentioning that our model focuses on the boundary of the document and most extracted phrases were located at the start of the document, which is controlled by our setting of $\alpha$. This proves the effectiveness of our boundary-aware centrality. 
From the figure, we also can find that wrong phrases are highly relevant to topics of this document, which is influenced by our phrase-document relevance weighting.
This example shows that the joint modeling of global and local context can improve the performance of keyphrase extraction and our model really captures local and global information.

\section{Related Work}
\subsection{Pre-trained Language Model}
Pre-trained language model is the kind of model that is trained on large-scale unlabeled corpus to learn prior knowledge and then fine-tuned on downstream tasks. The pre-trained language model without fine-tuning also can provide high quality embedding of natural texts for unsupervised tasks.
Different from static word embedding, such as Word2Vec \cite{mikolov-2013-word}, GloVe \cite{pennington-etal-2014-glove}, and FastText \cite{joulin-etal-2017-bag}. Pre-trained language models can encode words or sentences with context dynamically and solve the OOV problem.
In addition, pre-trained language models can provide document-level or sentence-level embedding which contains more semantic information than Sen2Vec \cite{pgj2017unsup} or Doc2Vec \cite{10.5555/3044805.3045025}.

ELMo \cite{peters-etal-2018-deep} employs Bi-LSTM structure and concatenate forward and backward information to capture bidirectional information. BERT \cite{devlin2019bert} is a bidirectional transformer structure pre-trained language model. Compared with the concatenation of bidirectional information, BERT can capture better context information. There are also a lot of other pre-trained language models such as RoBERTa \cite{liu2019roberta}, XLNET \cite{yang2020xlnet}, etc. In this paper, we choose BERT, the most used, to obtain vector representation of documents and phrases by merging the embedding of tokens.

\subsection{Unsupervised Keyphrase Extraction}
Unsupervised keyphrase extraction can be divided into four main types: statistics-based models, graph-based models, topic-based models, and embedding-based models. Statistics-based models \cite{yake-2018} mainly analyze an article’s probability features such as word frequency feature, position feature, linguistic features, etc. Topic-based models \cite{jardine-teufel-2014-topical,liu-etal-2009-clustering} focus on how to mine keyphrases by making use of the probability distribution of articles.

Graph-based models are the most proposed and popular used in early works which convert the document into a graph. Inspired by \cite{page1999pagerank}, \cite{mihalcea-2004-graph} proposed TextRank to convert keyphrase extraction task into the rank of nodes in graph.
After this, various works focused on the expansion of TextRank.
\cite{wan-2008-single} proposed SingleRank, which employs co-occurrences of tokens as edge weights. 
\cite{bougouin-etal-2013-topicrank} proposed TopicRank, which assigns a significance score to each topic by candidate keyphrase clustering. MultipartiteRank \cite{boudin-2018-unsupervised} encodes topical information within a multipartite graph structure.
Recently, \cite{Wang2015CorpusindependentGK} proposed WordAttractionRank, which added distance between word embeddings into SingleRank, and \cite{florescu-caragea-2017-positionrank} use node position weights, favoring words appearing earlier in the text. This position bias weighting strategy is very useful in news articles and long documents.

Embedding-based models benefit from the development of representation learning, which maps natural language into low-dimension vector representation.
Therefore, in recent years, embedding-based keyphrase extraction \cite{wang-etal-2016-extracting,bennani-smires-etal-2018-simple,papagiannopoulou2018local,ieee-sun-etal-2020-sifrank} has achieved good performance .
\cite{bennani-smires-etal-2018-simple} proposed EmbedRank, which ranks phrases by measuring the similarity between phrase embedding and document embedding. \cite{ieee-sun-etal-2020-sifrank} proposed SIFRank, which improves the static embedding from EmbedRank with a pre-trained language model.

Embedding-based models just measured the similarity between document and candidate phrases and ignored the local information. To jointly model global and local context \cite{zheng-lapata-2019-sentence,liang-etal-2021-improving}, in this paper, we revisit degree centrality, which can model local context, and convert it into boundary-aware centrality.
Then, we combine global similarity and boundary-aware centrality for local salient information to rank and extract phrases. 

\section{Conclusion and Future Work}
In this paper, we point out that embedding-based models ignore the local information and propose a novel model which jointly models global and local context. Our model revisited degree centrality and modified it with boundary function for modeling local context. We combine global similarity with our proposed boundary-aware centrality to extract keyphrases. Experiments on 3 public benchmarks demonstrate that our model can effectively capture global and local information and achieve remarkable results.
In the future work, we will focus on how to introduce our boundary-aware mechanism into supervised end2end keyphrase extraction/generation models.

\section*{Acknowledgments}
We thank the three anonymous reviewers for their careful reading of our paper and their many insightful comments and suggestions.
This work was supported in part by the National Natural Science Foundation of China (Grant Nos.U1636211, 61672081,61370126), the 2020 Tencent Wechat Rhino-Bird Focused Research Program, and the Fund of the State Key Laboratory of Software Development Environment (Grant No. SKLSDE-2021ZX-18). 

\bibliography{anthology,custom}
\bibliographystyle{acl_natbib}




\end{document}